\documentclass[12pt,onecolumn]{IEEEtran}
\usepackage[colorlinks,bookmarksopen,bookmarksnumbered,citecolor=red,urlcolor=red,]{hyperref}

\usepackage{color}
\usepackage{amsmath,bm} 

\usepackage{amsmath,mathtools}

\usepackage{verbatim}
\usepackage{epsfig,bbm}
\usepackage[colorlinks,bookmarksopen,bookmarksnumbered,citecolor=red,urlcolor=red,]{hyperref}
\usepackage{CJK}
\usepackage{indentfirst}
\usepackage{multirow}
 \usepackage{epstopdf}
\usepackage{graphicx}
\usepackage{footmisc}
\graphicspath{{./figures/}}
\usepackage{amsfonts}
\usepackage{mathrsfs}
\usepackage{setspace}
\usepackage{amsmath}
\usepackage{algorithm,algorithmic,amsbsy,amsmath,amssymb,epsfig,bbm,mathrsfs, bbm} 
\usepackage{amsthm}
\usepackage{verbatim} 
\usepackage[noadjust]{cite} 
\usepackage[subfigure]{graphfig}

\DeclarePairedDelimiterX{\infdivx}[2]{(}{)}{%
  #1\;\delimsize\|\;#2%
}
\newcommand{\infdiv}{D\infdivx}

\usepackage{geometry}
\usepackage{xcolor} 
\allowdisplaybreaks
\usepackage{xcolor} 
\allowdisplaybreaks

\mathtoolsset{showonlyrefs}

\def \xi {\mathbf{x}_i}

\usepackage{caption}

\begin{document}

\title{Detecting Cyberattacks in Industrial Control Systems Using Online Learning Algorithms}
\author{Guangxia Li, Yulong Shen, Peilin Zhao, Xiao Lu, Jia Liu, Yangyang Liu, Steven C. H. Hoi}

\maketitle

\begin{abstract}
Industrial control systems are critical to the operation of industrial facilities, especially for critical infrastructures, such as refineries, power grids, and transportation systems.
Similar to other information systems, a significant threat to industrial control systems is the attack from cyberspace---the offensive maneuvers launched by ``anonymous'' in the digital world that target computer-based assets with the goal of compromising a system's functions or probing for information.
Owing to the importance of industrial control systems, and the possibly devastating consequences of being attacked, significant endeavors have been attempted to secure industrial control systems from cyberattacks.
Among them are intrusion detection systems that serve as the first line of defense by monitoring and reporting potentially malicious activities.
Classical machine-learning-based intrusion detection methods usually generate prediction models by learning modest-sized training samples all at once.
Such approach is not always applicable to industrial control systems, as industrial control systems must process continuous control commands with limited computational resources in a nonstop way.
To satisfy such requirements, we propose using online learning to learn prediction models from the controlling data stream.
We introduce several state-of-the-art online learning algorithms categorically, and illustrate their efficacies on two typically used testbeds---power system and gas pipeline.
Further, we explore a new cost-sensitive online learning algorithm to solve the class-imbalance problem that is pervasive in industrial intrusion detection systems.
Our experimental results indicate that the proposed algorithm can achieve an overall improvement in the detection rate of cyberattacks in industrial control systems.
\end{abstract}

\section{Introduction}
\label{sec1}

Modern industrial control systems are microprocessor-equipped devices and associated communication networks used to monitor and operate physical equipment in the industrial environment.
Such systems are designated to collect sensor measurements and operational data from the physical world, display information to human operators, perform decisions based upon the detected events, and issue control commands to the controlled equipment.
The commands are used to turn on or off power switches, open or close hydraulic valves, adjust motor speed, shut down engines in emergencies, etc.
Although such operations are routine, they are crucial in industrial processes as any misoperation can cause incidents that may lead to devastating consequences in terms of financial loss, acute health effects, or environmental impacts.
Modern digit-controller-based industrial control systems exhibit many advantages compared to their predecessors such as mechanical-based and electromechanical-based systems in terms of performance, reliability, and cost.
In fact, modern industrial control systems have been applied widely in practice, and are the de-facto standard configuration of almost every industrial sector.
\begin{figure}[th]
\centering
\includegraphics[width=1.0\textwidth]{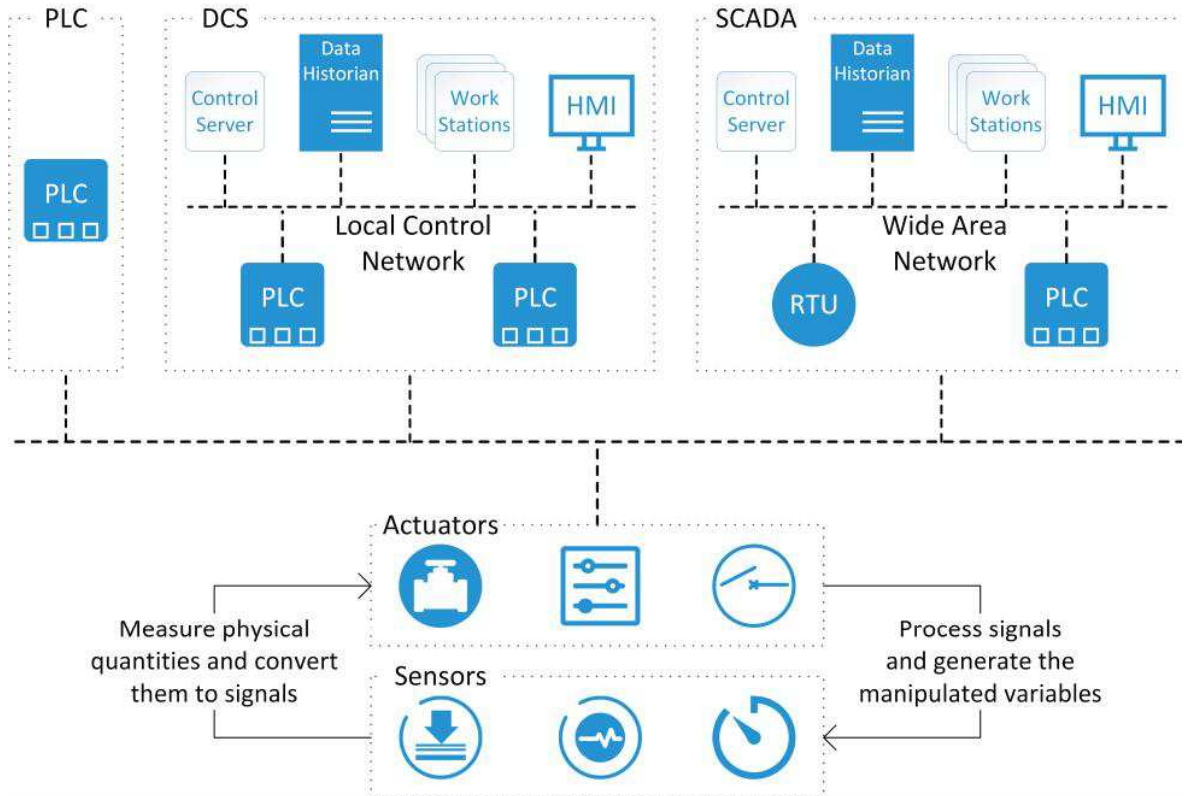}
\caption{Notional topology of an industrial control system.}
\label{fig1-1}
\end{figure}

Figure~\ref{fig1-1} shows a notional topology of an industrial control system.
As shown, the sensors measure physical quantities (e.g., flow, pressure, speed) and convert them into signals that are transmitted to the controllers.
The controllers process the sensor signals to generate manipulated variables that are sent to the actuators (e.g., breakers, switches, valves) to manipulate the controlled process directly.
Sensors, actuators, and controllers, together with some external components such as human machine interfaces (HMIs) and remote maintenance tools compose a typical industrial control system.
Regarding the actual implementation of the system, many variants exist and their boundaries can be blurry.
Still, there are several types of widely used control systems, such as supervisory control and data acquisition (SCADA) systems \cite{niyato2011machine}, distributed control systems (DCSs) \cite{li2019data}, and programmable logic controllers (PLCs).
Specifically, a SCADA system comprises a control center, and one or more geographically distributed field sites consisting of PLCs and/or remote terminal units (RTUs) used to command actuators and sensors. 
It is generally used to control geographically dispersed assets.
A DCS, by contrast, is always applied to control production systems within a local area using the supervisory and regulatory control mechanism.
As for PLCs, except for serving as the local controllers in SCADA and DCS configurations, they can also be implemented as the primary controllers in some smaller control systems to provide closed-loop control with no direct human involvement.
For details about SCADA, DCS, PLC, and other types of control systems, refer to Stouffer et al.~\cite{stouffer2011guide}.
 
Industrial control systems are critical to the operation of industrial facilities, particularly to national critical infrastructures, such as refineries, chemical plants, electrical power grids, oil and natural gas pipelines, and transportation systems.
Their incidents can cause significant risk to human lives and serious damage to the environment.
Industrial control systems had been thought to be immune to outsider threats because they were originally designed as isolated systems running proprietary control protocols using specialized hardware and software.
This could be true in the past, but is no longer applicable nowadays.
Modern industrial control systems do not operate in isolation anymore, but tend to be connected to wider networks (e.g., Internet, Internet-of-things~\cite{niyato2015economics}, sensor networks \cite{niyato2016distributed,lu2016sensor}, smart grid systems \cite{LuXiaoPower,niyato2012adaptive,korki2011mac}, cloud systems \cite{li2019data}, communication systems \cite{lu2019intelligent,niyato2015game,lu2019coverage,lu2018ambient}, corporate networks \cite{lu2011payoff,lu2015hierarchical,lu2019ambient,lu2018performance}, and mobile social networks~\cite{zhang2015optimizing}).
The proprietary protocols once unfamiliar to the public are gradually being replaced by open standards such as the Ethernet, TCP/IP \cite{lu2010sender,lu2014ssthreshless}, and web services.
The merging of typical information technologies into industrial control systems reduces the dubious protective barrier of ``security by obscurity'', and thus increases the possibility of cybersecurity vulnerabilities and incidents \cite{lu2018cyber,lu2018managing,lu2018cyber2,niyato2015performance}.
In fact, cyberattacks to industrial control systems have occurred at an alarming pace in the last decade.
Recent records include Stuxnet, Davis-Besse Nuclear Plant, Maroochy in Australia, Flame, and Aurora~\cite{stouffer2011guide}.

Among the recent cyberattacks is the famous Stuxnet worm known for its unverified but highly possible intent to compromise Iran's nuclear program.
Uncovered in 2010, Stuxnet is the first identified malware that targets SCADA systems~\cite{falliere2011w32}.
It is believed to be introduced to Iran's industrial sites via an infected USB flash drive.
Subsequently, it propagates across the network using Microsoft Windows flaws.
Stuxnet's spread is indiscriminate, but its attack is designated to target only Siemens S7-300 PLC systems with particular variable frequency drives (VFD) attached.
In particular, it monitors the frequency of motors controlled by VFD, and only attacks systems that spin between 807 Hz and 1210 Hz.
The industrial applications of motors with these parameters include high-speed centrifuges that are essential for uranium enrichment.
Stuxnet periodically modifies the frequency of VFD, and thus causes the rotational speed of connected motors to change in an unusual manner.
It fakes the sensor signal to monitor systems, rendering the deed unbeknownst to human operators.
Consequently, the fast-spinning centrifuges become destabilized initially and finally break down.
Such sophisticated plots, together with the unprecedented complexity of the code, strongly suggest that Stuxnet is not a hacker's sabotage, but a state-sponsored cyberattack.
Because Stuxnet's design is not domain specific, it could be tailored as a platform for attacking any SCADA or PLC system.
This is proven by a number of new worms found subsequently, considered to be related to Stuxnet.

The recognition of Stuxnet has intensified the public's awareness of the information security of industrial control systems.
However, securing them is not easy.
Owing to their long life span, legacy systems with great vulnerability are still active currently.
It is not unusual for outdated devices without security patch to be manipulated by unalerted technicians in industrial sites.
Because of the real-time, continuity, and constrained environment of the industrial control system, many methods used in traditional computer security (such as virus database update) are difficult to apply.
As incidents and malicious actions are inevitable, detecting their occurrences timely is important to industrial control system administrators, and automatic devices such as firewalls and anti-virus software.
This can be achieved by adapting solutions for traditional information technology environments to develop industrial-control-system-specific intrusion detection systems.

Artificial-intelligence-based approaches such as machine learning have been employed widely by intrusion detection systems.
Among various machine learning methods, online learning represents a family of efficient algorithms that can build the predictor incrementally by processing the training data in a sequential manner, as opposed to batch learning algorithms that train the predictor by learning the entire dataset all at once~\cite{DBLP:journals/corr/abs-1802-02871}.
Specifically, online learning algorithms perform on a sequence of data by processing them one by one.
On each round, the learner receives an input, makes a prediction using an internal hypothesis that is retained in memory, and subsequently learns the true label.
It uses the new example to modify its hypothesis according to some predefined rules.
The goal is to minimize the total number of rounds with incorrect predictions.
In general, online learning algorithms are fast, simple, and require few statistical assumptions, rendering them applicable to a wide range of applications.
They can scale well to a large amount of data, and are particularly suitable for real-world applications where data arrive continuously.

We herein study the problem of detecting cyberattacks in industrial control systems using online learning algorithms.
We evaluate several state-of-the-art online learners in terms of their ability to identify malicious control commands from normal ones using testbeds provided by the Critical Infrastructure Protection Center of the Mississippi State University.
The experimental results on a power system and a gas pipeline testbed indicate that online learning algorithms can discriminate the intrusions effectively.
Furthermore, we focus on the so-called class-imbalanced problem that troubles most intrusion detection systems in real-world applications: relatively large numbers of normal events exist that can easily overwhelm a few attacks and distract classifiers.
To address the class-imbalanced problem, we propose a cost-sensitive online learning method, namely the adaptive regularized cost-sensitive multiclass online learning, such that the classifier can focus on the minority classes that are more important.
The proposed algorithm is a combination of the second-order online learning technique and the cost-sensitive learning approach.
It differs from traditional multiclass online learners who are only concerned about the performance in terms of prediction mistake rate by taking the misclassification costs into consideration.
We demonstrate experimentally that the proposed algorithm can discriminate attacks precisely and efficiently.  
 
The remainder of this paper is organized as follows.
Section~\ref{sec2} reviews the related work on industrial control system security.
Section~\ref{sec3} begins with an introduction to classical online learning algorithms.
Subsequently, our adaptive regularized cost-sensitive multiclass online learning algorithm is presented.
Section~\ref{sec4} gives experimental results and discussion.
Section~\ref{sec5} concludes this paper.
 
\section{Related Work}
\label{sec2}

In recent years, industrial control system security has garnered increasing attention, particularly that pertaining to critical infrastructures.
Several academic studies have been performed to understand the characteristics of industrial control systems, analyze their vulnerabilities to cyberattacks, simulate real systems with testbeds, and demonstrate the importance of cybersecurity in industrial control systems.
C{\'{a}}rdenas et al.~\cite{DBLP:conf/uss/CardenasAS08} discussed the differences between the security of control systems and traditional IT systems, and analyzed the reasons for the current control systems being more vulnerable than before to cyberattacks.
Companies, organizations, and government agencies are involved in the industrial control system security initiatives as well, primarily in the form of publishing guidelines, standards, and best practices.
As an example, the U.S. National Institute of Standards and Technology (NIST) published a number of guidances to cybersecurity risk management, among which is the NIST Special Publication 800-82 providing cross-industry guidelines for securing industrial control systems~\cite{stouffer2011guide}.
This publication highlights the typical threats and vulnerabilities to these systems and provides the recommended the security countermeasures to mitigate the associated risks.
In addition to the guidance to cross industry, there are a number of publications targeting specific industries such as chemical, oil and gas.
The surveys of such guidance, together with methodologies for measuring and managing threats, can be found in~\cite{DBLP:journals/ijcip/KnowlesPHDJ15,DBLP:journals/ijon/DingHXGZ18}.

Most of the conventional methods for protecting control systems have focused on increasing their reliability and maintainability.
However, an urgent growing concern has emerged for protecting control systems against attacks launched in cyberspace.
To detect such attacks, one can rely on the profile of anomaly patterns~\cite{DBLP:journals/tii/CarcanoCGMFT11,DBLP:journals/tii/PanMA15,DBLP:journals/tsg/PanMA15,DBLP:journals/jdfsl/GaoM14}.
As an example, Carcano et al.~\cite{DBLP:journals/tii/CarcanoCGMFT11} proposed an intrusion detection method for SCADA system based on tracking the so-called critical states that correspond to dangerous or unwanted situations in the monitored system.
Their approach assumed that cyberattacks are always performed by forcing a transition of the system from a safe state to a critical state.
Because the critical states are generally well known and limited in number, one can enumerate them formally beforehand and predict the criticality by tracking the changes in the distance between the current system state and the critical states.
Likewise, Pan et al.~\cite{DBLP:journals/tii/PanMA15} processed a sequence of critical system states using a sequential pattern mining algorithm to detect disturbances and cyberattacks in power systems.
In contrast to exploiting the abnormal patterns, one can also specify the acceptable behaviors of a system, and subsequently detect attacks that cause violation to them~\cite{cheung2007using,DBLP:journals/tsg/MitchellC13}.
The so-called specification-based intrusion detection approach monitors the system according to policies specified by valid sequences of system behaviors.
Any sequence of behaviors outside the predefined specifications is regarded as an abnormal behavior.
As a representative work, Cheung et al.~\cite{cheung2007using} presented three model-based detection techniques for monitoring SCADA networks: (1) specifying the expected characteristics of network request/response according to the Modbus protocol; (2) defining the expected communication patterns among network components; and (3) detecting the changes in server/service availability.

The aforementioned rule-based intrusion detection methods generally rely on human efforts to transform expert knowledge into machine-executable rules.
Manually constructing such rules, however, can be a laborious and expensive endeavor.
Machine learning based methods prevail in this situation as they can automatically generate rules from the existing examples without human efforts.
For performance testing, Beaver et al.~\cite{DBLP:conf/icmla/BeaverBB13} evaluated several classical machine learning algorithms including the decision tree, na\"{i}ve Bayes classifier, and support vector machine (SVM) in terms of their ability to identify cyberattacks using a dataset of RTU communications in a gas pipeline system.
A similar evaluation was performed on a power system to demonstrate the feasibility of applying machine learning algorithms to discriminate types of power system disturbances, especially those caused by malicious attacks~\cite{hink2014machine}.
Terai et al.~\cite{DBLP:conf/eurosp/TeraiAKTK17} built a discriminant model between a normal operation and attack packets on a laboratorial fluid system equipped with an actual industrial controller using SVM with packet transmission intervals and length as features.
Schuster et al.~\cite{DBLP:conf/ssci/SchusterPRK15} applied the one-class classification technique to implement a self-configuring anomaly detection for industrial network data.
They identified the one-class SVM as a promising learning method for this task, as no sample of attacks or other anomalous traffic is required to construct the training set.
Other classical intrusion detection methods employing machine learning techniques include exemplar-based classifier (e.g.,~\cite{DBLP:journals/tsg/AdhikariMP18}), k-nearest neighbors (e.g.,~\cite{DBLP:journals/compsec/LiaoV02,DBLP:conf/raid/WangS04}), neural network (e.g.,~\cite{mukkamala2002intrusion,DBLP:journals/cor/ChenHS05}), SVM (e.g.,~\cite{mukkamala2002intrusion,DBLP:journals/cor/ChenHS05,DBLP:journals/jnca/MukkamalaSA05}), and na\"{i}ve Bayes (e.g.,~\cite{DBLP:journals/eswa/KocMS12}).
Sommer et al.~\cite{DBLP:conf/sp/SommerP10} identified a few unique challenges and corresponding guidelines regarding the use of machine learning for anomaly detection in a general network environment.
Mantere et al.~\cite{DBLP:conf/aina/MantereUSN12} narrowed down the scope from general networks to industrial control systems, and argued that the diversity of network traffic, while prevails in general networks and tends to disturb machine learning algorithms, is decreased significantly in industrial control systems.
They thus considered machine learning a promising tool for intrusion detection in industrial control systems.

\section{Online Learning for Intrusion Detection}
\label{sec3}

\subsection{Problem Setting}
\label{sec:3.1}

We tackle the intrusion detection problem with supervised machine learning techniques.
Specifically, the task of detecting malicious actions from normal actions can be cast as a binary classification problem, in which we use a positive class to denote the malicious actions, and a negative class for the normal actions.
To further distinguish malicious action types, the task can be transformed into a multiclass classification problem. 
To learn a classifier with machine learning techniques, a training set consisting of samples whose class labels are known is required.
The training set consisting of samples of attributes and associated class labels is used to build a classification model, which is subsequently applied to samples with unknown class labels.
A learning algorithm is employed to build a model that estimates the relationship between the attributes and class label of the training data.
The model generated by a learning algorithm should both fit the input data well and correctly predict the class labels of samples that it has never seen.

Several well-known machine learning algorithms such as the decision tree, neural network, and SVM belong to the batch learning paradigm which assumes that all training samples are available before the learning process occurs.
In contrast to batch learning, online learning algorithms operate on a stream of data by deciding the present instance based on past knowledge together with the latest available information.
Formally, at each step $t$, the learner receives an incoming sample $(\mathbf{x}_t, y_t)$, where $\mathbf{x}_t \in \mathbb{R}^d$ is a $d$-dimensional vector representing the data, and $y_t$ refers to its class label.
For binary classification, $y_t \in \{-1,1\}$; for multiclass classification, $y_t \in \{1,\ldots,k\}$ and $k \geq 3$.
The classification model to learn is parameterized by a weight vector $\mathbf{w} \in \mathbb{R}^d$.
The learner first predicts the class label of the incoming instance as $\hat{y}_t$ according to some criterion $\sigma(\mathbf{w} \cdot \mathbf{x})$.
After the prediction, the true label $y_t$ is revealed.
It subsequently computes the loss $\ell(\hat{y}_t, y_t)$ according to the difference between the prediction $\hat{y}_t$ and the revealed true label $y_t$, and updates the model by a certain strategy.
The goal is to learn a model to minimize the online regret measured as the difference between the cumulative loss of the online learning algorithm and the cumulative loss of the best model, i.e., $\sum^T_{t=1}\ell(\sigma(\mathbf{w}_t \cdot \mathbf{x}_t), y_t) - \min_\mathbf{w} \sum^T_{t=1}\ell(\sigma(\mathbf{w} \cdot \mathbf{x}_t), y_t)$.
Different updating strategies lead to different online learning algorithms.
We elaborate some representative algorithms as follows.

\subsection{Online Binary Classification}
\label{sec:3.2}

The seminal work of Frank Rosenblatt~\cite{rosenblatt1958perceptron} proposed a simple model called perceptron.
The perceptron operates by assigning weights to incoming connections.
At each learning round, it takes the dot-product of each incoming value with a weight, and subsequently verifies if it is over or below a certain threshold.
It compares the predicted label $\hat{y}_t = \mathrm{sign}(\mathbf{w}_t \cdot \mathbf{x}_t)$ with the true label $y_t \in \{-1,1\}$.
If $\hat{y}_t \neq y_t$, the perceptron updates the model as $\mathbf{w}_{t+1} = \mathbf{w}_t + y_t \mathbf{x}_t$.

As improvements to perceptron-like algorithms, many modern online learning methods~\cite{DBLP:journals/jmlr/Gentile01,DBLP:journals/ml/LiL02,DBLP:conf/icml/Zinkevich03,DBLP:journals/jmlr/CrammerDKSS06} have been proposed over the past decades.
They are partly inspired by the maximum margin learning principle that has been applied successfully to batch mode learning.
Specifically, for the incoming example $(\mathbf{x}_t, y_t)$ and the algorithm's weight vector $\mathbf{w}_t$, the term $y_t (\mathbf{w}_t \cdot \mathbf{x}_t)$ is referred to as the (signed) margin.
Whenever the margin is a positive number, we say that the algorithm has predicted correctly.
However, we are not satisfied with a positive margin value; we would prefer for the algorithm to predict correctly and with a larger margin.
Therefore, our goal is to achieve a margin of at least 1, as often as possible.
On rounds where the algorithm attains a margin less than 1, it suffers an instantaneous loss.
Typically, this loss is defined by the following hinge-loss function:
\begin{equation}
\ell(\mathbf{w}; \mathbf{x}_t, y_t) = \max (0, \; 1-y_t (\mathbf{w} \cdot \mathbf{x}_t))
\label{eq3.2-1}
\end{equation}

An example of such approach is the passive-aggressive (PA) algorithm~\cite{DBLP:journals/jmlr/CrammerDKSS06}.
In addition to employing the maximum margin principle, PA maintains a trade-off between the amount of progress achieved on each training round and the information retained from the previous rounds.
On one hand, the classifier should be updated whenever it misclassifies a new instance.
On the other hand, the classifier should not be changed too rapidly especially if it predicts most of the previous instances correctly.
Formally, it is formulated as the following optimization problem:
\begin{align}
\mathbf{w}_{t+1} & = \displaystyle{\mathop{\mathrm{argmin}}_{\mathbf{w} \in \mathbb{R}^d}} \;
\frac{1}{2}{\left\| \mathbf{w} - \mathbf{w}_t \right\|}^2 \\
\mbox{s.t.} & \quad \ell(\mathbf{w}; \mathbf{x}_t, y_t) = 0
\label{eq3.2-2}
\end{align}
where $\ell(\mathbf{w}; \mathbf{x}_t, y_t)$ is the hinge loss defined in Eq.~\eqref{eq3.2-1}.

In addition, some variants of PA are proposed to use the soft-margin technique to handle the non-separable and noisy cases. 
As an example, a variant named PA-I is formulated as follows:
\begin{align}
\mathbf{w}_{t+1} & = \displaystyle{\mathop{\mathrm{argmin}}_{\mathbf{w} \in \mathbb{R}^d}} \;
\frac{1}{2}{\left\| \mathbf{w} - \mathbf{w}_t \right\|}^2 +
C \xi \\
\mbox{s.t.} & \quad \ell(\mathbf{w}; \mathbf{x}_t, y_t) \leq \xi \quad \mbox{and} \quad \xi \geq 0  
\label{eq3.2-3}
\end{align}
where $C$ is a positive parameter that controls the influence of the slack term $\xi$ on the objective function.

The online learning algorithms above belong to the family of first-order methods, as they only depend on the first-order information of the example.
Additionally, the machine learning community has studied the second-order online learning algorithms that use parameter confidence information to guide the learning process.
A family of confidence-weighted learning algorithms~\cite{DBLP:conf/icml/DredzeCP08,CrammerDP08,DBLP:conf/nips/CrammerKD09} assumes that the weight vector follows a Gaussian distribution $\mathbf{w} \sim \mathcal{N}(\mathbf{\mu},\Sigma)$ with mean vector $\mathbf{\mu} \in \mathbb{R}^d$ and covariance matrix $\Sigma \in \mathbb{R}^{d \times d}$.
To classify an instance $\mathbf{x}$, it draws a parameter vector $\mathbf{w} \sim \mathcal{N}(\mathbf{\mu},\Sigma)$ and predicts the label according to $\mathrm{sign}(\mathbf{w} \cdot \mathbf{x})$.
In practice, however, the average weight vector $E(\mathbf{w}) = \mathbf{\mu}$ is used for the prediction.
The model parameters, including both $\mathbf{\mu}$ and $\Sigma$ are updated appropriately with the effect of controlling the direction and scale of parameter updates.
The learner performs online updates based on its confidence in the current parameters, generating larger changes in the weights of infrequently observed features.
Our empirical evaluation in the following indicates the advantages of learning with the second-order information.

As an example, the adaptive regularization of weight vectors (AROW) algorithm~\cite{DBLP:conf/nips/CrammerKD09} maintains a probabilistic measure of confidence in each component of its weight vector using a Gaussian distribution.
The weight distribution is updated by minimizing the Kullback--Leibler (KL) divergence between the new and old weight distributions under the constraint that the probability of correct classification is greater than a threshold.
At round $t$, when receiving $(\mathbf{x}_t, y_t)$, the model is updated by minimizing the following objective:
\begin{align}
\displaystyle{\mathop{\mathrm{argmin}}_{\mathbf{\mu} \in \mathbb{R}^{d}, \Sigma \in \mathbb{R}^{d \times d}}} 
D_{KL} \infdiv{\mathcal{N}(\mathbf{\mu},\Sigma)}{\mathcal{N}(\mathbf{\mu}_t,\Sigma_t)} + 
\frac{1}{2\gamma}{\ell^2 ({\mathbf{\mu}}; (\mathbf{x}_t,{y_t}))} +
\frac{1}{2\gamma}{{\mathbf{x}}_t^{\mathrm{T}} \Sigma {\mathbf{x}}_t} 
\label{eq3.2-4}
\end{align}
where $D_{KL} \infdiv{\mathcal{N}(\mathbf{\mu},\Sigma)}{\mathcal{N}(\mathbf{\mu}_t,\Sigma_t)}$ is the KL divergence.

The above minimization problem can be solved with a closed-form solution as in~\cite{DBLP:conf/nips/CrammerKD09}.
This makes AROW quite fast as it does not need to invoke any optimization routine for updating.

\subsection{Online Multiclass Classification}
\label{sec:3.3}

The intrusion detection task occasionally requires the discrimination of attack type, instead of only distinguishing attacks from normal actions.
Compared to the aforementioned binary classification, online multiclass classification operates over the same sequence of data samples $(\mathbf{x}_1, y_1), \ldots ,(\mathbf{x}_T, y_T)$, but differs in that the choice of labels has more than two odds, i.e., $y_t \in \{1, \ldots, k\}$ and $k \geq 3$. 
Unlike binary classification that represents the model with a weight vector $\mathbf{w} \in \mathbb{R}^d$, the multiclass model contains a $k \times d$ matrix $\mathbf{W}$, whose \textit{i}th row can be considered as the model for the \textit{i}th class.
Specifically, the compound weight $\mathbf{W}$ is composed as follows:
\begin{equation}
\mathbf{W} =
\begin{bmatrix}
\mathbf{w}_1 \\
\mathbf{w}_2 \\
\vdots \\
\mathbf{w}_k
\end{bmatrix}
\label{eq3.3-1}
\end{equation}

Given an input $\mathbf{x}_t$, the online multiclass algorithm predicts the label as the index associated with the largest prediction value, i.e.,
\begin{equation*}
\hat{y}_t = \displaystyle{\mathop{\mathrm{argmax}}_{i \in [k]}} \;
\mathbf{W}_{t,i} \cdot \mathbf{x}_t
\label{eq3.3-2}
\end{equation*}
where $\mathbf{W}_{t,i}$ is the \textit{i}th row of the matrix $\mathbf{W}$ as shown in Eq.~\eqref{eq3.3-1}.

After generating a prediction, the learner computes the loss based on the true label $y_t$ and the highest-ranked irrelevant label as follows:
\begin{equation*}
\ell_{mc}(\mathbf{W}_t; \mathbf{x}_t, y_t) = 
\max(0, \; 1 - (\mathbf{W}_{t,y_t} \cdot \mathbf{x}_t - \max_{i \in [k], i \neq y_t} \mathbf{W}_{t,i} \cdot \mathbf{x}_t))
\label{eq3.3-3}
\end{equation*}
The learner subsequently updates the model parameter $\mathbf{W}_t$ based on certain rules as those described in~\cite{DBLP:journals/jmlr/CrammerS03,DBLP:conf/icml/FinkSSU06}.

\subsection{Cost-Sensitive Online Learning for Binary Classification}
\label{sec:3.4}

A significant trait of the intrusion detection task is the skewed distribution of classes.
In most cases, most of the classes are normal events.
The misclassification costs of instances from different classes can be significantly different.
The classical online learning algorithms minimize the regret, or equally, maximize the accuracy.
However, pursuing a maximal accuracy may be inappropriate on imbalanced datasets because a trivial learner that simply classifies all samples as negative could achieve a high accuracy, but is of little use in practice.
This renders it unsuitable for class-imbalanced datasets.

As a remedy for class-imbalanced datasets, the cost-sensitive classification differs from the normal classification approach by considering the misclassification costs during the training process~\cite{DBLP:conf/ijcai/Elkan01}.
In principle, a better performance might be obtained if the classifier is tailored by the learning algorithm using the cost matrix.
Over the past decades, substantial research efforts have been devoted to developing cost-sensitive classification algorithms.
In online learning, cost-sensitive classification methods exist for the binary class case~\cite{DBLP:journals/jmlr/CrammerS03,DBLP:conf/icml/FinkSSU06,DBLP:conf/icdm/ZhaoZWLH15}.
The key is to change the hinge loss in Eq.~\eqref{eq3.2-1} to incorporate cost-sensitive measures and optimize such measures directly.

As an example, the cost sensitive online gradient descent (CSOGD) algorithm~\cite{DBLP:conf/icdm/ZhaoZWLH15} is proposed to maximize the sum of weighted sensitivity (the proportion of positive samples that are identified correctly as such) and specificity (the proportion of negatives that are identified correctly as such).
Hence, it modifies the hinge loss function as follows:
\begin{equation}
\ell(\mathbf{w}; \mathbf{x}_t, y_t) = 
\max (0, \; (\rho \ast \mathrm{I}_{y=1} + \mathrm{I}_{y=-1})-y_t (\mathbf{w} \cdot \mathbf{x}_t))
\label{eq3.4-1}
\end{equation}
where $\rho$ is a predefined parameter related to the ratio of the number of negative samples to the number of positive samples, and $\mathrm{I}$ is an indicator function.

The model is updated by:
\begin{equation*}
\mathbf{w}_{t+1} = \mathbf{w}_t - \lambda \nabla \ell(\mathbf{w}_t)
\label{eq3.4-2}
\end{equation*}
where $\lambda$ is a learning rate parameter and $\nabla \ell(\mathbf{w}_t)$ is the gradient of loss function in Eq.~\eqref{eq3.4-1}

\subsection{Cost-Sensitive Online Learning for Multiclass Classification}
\label{sec:3.5}

Despite being studied extensively for binary classification problems, cost-sensitive online learning has been rarely examined for the multiclass case, even though the imbalanced class distribution prevails in real-world applications.
This particularly applies to industrial control systems where the attacks, with various subclasses, comprise only a small part of all events.
As the sensitivity, specificity, and other class-sensitive metrics are defined for binary classification, the aforementioned cost-sensitive learning technique cannot be applied to the multiclass case.
We thus propose a cost-sensitive online learning algorithm that can solve the multiclass classification problem.

Suppose that there are $k$ classes.
We use a $k \times d$ matrix $\mathbf{W}$ as defined in Eq.~\eqref{eq3.3-1} to represent the model.
To define the cost of misclassification, we use a $k \times k$ matrix $C$ in which the diagonal elements represent the cost of correct prediction (they are set to zero), and the off-diagonal elements $c(i,j) > 0, i, j \in \{1,2,\ldots,k\}, i \neq j$ denote the cost of misclassifying a sample of the \textit{i}th class to the \textit{j}th class.
Given an example $\mathbf{x}$, we define the most possible misclassified class as follows:
\begin{equation*}
p = \displaystyle{\mathop{\mathrm{argmax}}_{i \in [k], i \neq y_t}} \;
\mathbf{W}_{t,i} \cdot \mathbf{x}_t
\label{eq3.5-1}
\end{equation*}

The loss on this example is defined as follows:
\begin{equation}
\ell_{mc}(\mathbf{W}_t; \mathbf{x}_t, y_t)) = 
\max(0, \; c(y_t, p) - (\mathbf{W}_{t,y_t} \cdot \mathbf{x}_t - \mathbf{W}_{t,p} \cdot \mathbf{x}_t))
\label{eq3.5-2}
\end{equation}
where $c(y_t, p)$ is an element extracted from the predefined cost matrix $C$.

Replacing the loss function of any online learning algorithms with Eq.~\eqref{eq3.5-2} leads to a multiclass cost-sensitive online learning algorithm.
We specifically employ the adaptive regularization of weights (AROW)~\cite{DBLP:conf/nips/CrammerKD09} as the framework to derive a new algorithm, namely the adaptive regularized cost-sensitive multiclass online learning (ARCSMC).
This is summarized in Algorithm~\ref{alg1}.
The evaluation of the ARCSMC algorithm is presented in the next section.
\begin{algorithm}[t]
	\caption{Adaptive regularized cost-sensitive multiclass online learning (ARCSMC).}
	\label{alg1}
	\begin{algorithmic}[1]
		\STATE {\bf Input}: A sequence of samples $(\mathbf{x}_t, y_t)$ where $\mathbf{x}_t \in \mathbb{R}^{d}$, $y_t \in \{1,\ldots,k\}$ and $t \in \{1,\ldots,T\}$; a cost matrix $C$; a parameter $\gamma > 0$
		\STATE {\bf Output}: A vector $\mathbf{W} \in \mathbb{R}^{k \times d}$ representing the compound model
		\STATE {\bf Initialize}: Set $\mathbf{W}_0 = \mathbf{0}$, $\Sigma_0 = I$
		\FOR {$t=1,\ldots,T$}
		\STATE Receive an instance $\mathbf{x}_t$
		\STATE Make the prediction as $\hat{y}_t = \mathop{\mathrm{argmax}}_{i \in [k]} \mathbf{W}_{t,i} \cdot \mathbf{x}_t$
		\STATE Receive the true label $y_t$
		\STATE Compute the most possible misclassified class as $p =\mathop{\mathrm{argmax}}_{i \in [k], i \neq y_t} \mathbf{W}_{t,i} \cdot \mathbf{x}_t$
		\STATE Compute the loss as $\ell_t = \max(0, \; c(y_t, p) - (\mathbf{W}_{t,y_t} \cdot \mathbf{x}_t - \mathbf{W}_{t,p} \cdot \mathbf{x}_t))$
		\IF {$\ell_t > 0$}
        	\STATE Update the model as $\mathbf{W}_{t+1,y_t} = \mathbf{W}_{t,y_t} + \alpha_t \Sigma_t y_t \mathbf{x}_t$, and $\mathbf{W}_{t+1,p} = \mathbf{W}_{t,p} - \alpha_t \Sigma_t y_t \mathbf{x}_t$, where $\Sigma_{t+1} = \Sigma_t - \beta_t \Sigma_t \mathbf{x}_t \mathbf{x}_t^{\top} \Sigma_t$, $\alpha_t = \ell_t \beta_t$, and $\beta_t = \frac{1}{\mathbf{x}_t^{\top} \Sigma_t \mathbf{x}_t + \gamma}$
		\ENDIF
		\ENDFOR
	\end{algorithmic}
\end{algorithm}

\section{Experimental Results}
\label{sec4}

It is difficult to conduct the security experiment on real industrial control systems because of the potential risk and downtime of services provided by the facilities controlled by them.
An alternative method is to simulate their functions in an isolated environment, also known as the testbed, where experiments can be performed safely.
In our experiment, we used two testbeds to evaluate the performance of online learning algorithms.
We start by introducing our experimental setup, followed by discussion on the results.

\subsection{Experimental Testbeds}
\label{sec:4.1}

The data used in our experiment are extracted from two testbeds developed by the Mississippi State University's Critical Infrastructure Protection Center.

\paragraph{Power System Dataset}
The modern power transmission system, also known as the smart grid, relies on field sensors such as synchrophasors for remote monitoring and controlling.
The synchrophasor data contain measurements such as voltage and current phasor, as well as the status of system devices including relays, breakers, and transformers.
It is typically sampled at a high speed (e.g., 120 times per second~\cite{DBLP:journals/tsg/PanMA15}) and sent to a processing unit with low latency.
Such a configuration causes the system to generate a large volume of data that demands real-time processing---an ideal scenario for applying online learning algorithms.

We used a testbed to explore the suitability of applying online learning methods to discriminate malicious activities from natural power system disturbances.
The dataset includes the simulation of 37 event scenarios including natural disturbances (8 events), normal operations (1 event), and cyberattacks (28 events) in a two-line three-bus power system~\cite{hink2014machine}.
Two classification schemes are employed in the experiment: one is a binary classification where the 37 event scenarios are grouped as either the attack (28 events) or normal operation (9 events); the other is a three-class classification sharing the same setting as mentioned above.
There are 78,377 samples in the dataset.
Each sample consists of 128 features: 116 measurements are generated by 4 synchrophasors, and 12 measurements are from the control panel logs, relay logs, and Snort alerts (Snort is a network monitoring tool).
The sample size, feature count, and class distributions of this dataset are summarized in Table~1.

\paragraph{Gas Pipeline Dataset}
This dataset is a collection of labeled command/response streams from a simulated control system that models a gas pipeline used to transfer natural gas or other petroleum products~\cite{DBLP:journals/ijcip/MorrisSRGPR11,morris2015industrial}.
The physical system comprises a closed-loop gas pipeline connected to an air pump that pumps air into the pipeline, a manual release valve together with a solenoid release valve used to release air pressure from the pipeline, and a pressure sensor.
Commercial PLC, RTU, and HMI are configured to control the physical system to maintain a specific pipeline pressure value.

Artifacts of normal operations and cyberattacks are mixed randomly to compose the dataset.
Four categories of cyberattacks are included: response injection, reconnaissance, denial of service, and command injection.
The configuration details can be found in~\cite{DBLP:journals/ijcip/MorrisSRGPR11}.
Unlike the setting above, our experiment on this dataset only involves a binary classification task.
It consists of 274,628 samples, in which 60,048 samples are attack related.
Table~2 provides the statistics of this dataset.

\begin{figure}[tpb]
\centering 
\includegraphics[width=0.9\textwidth]{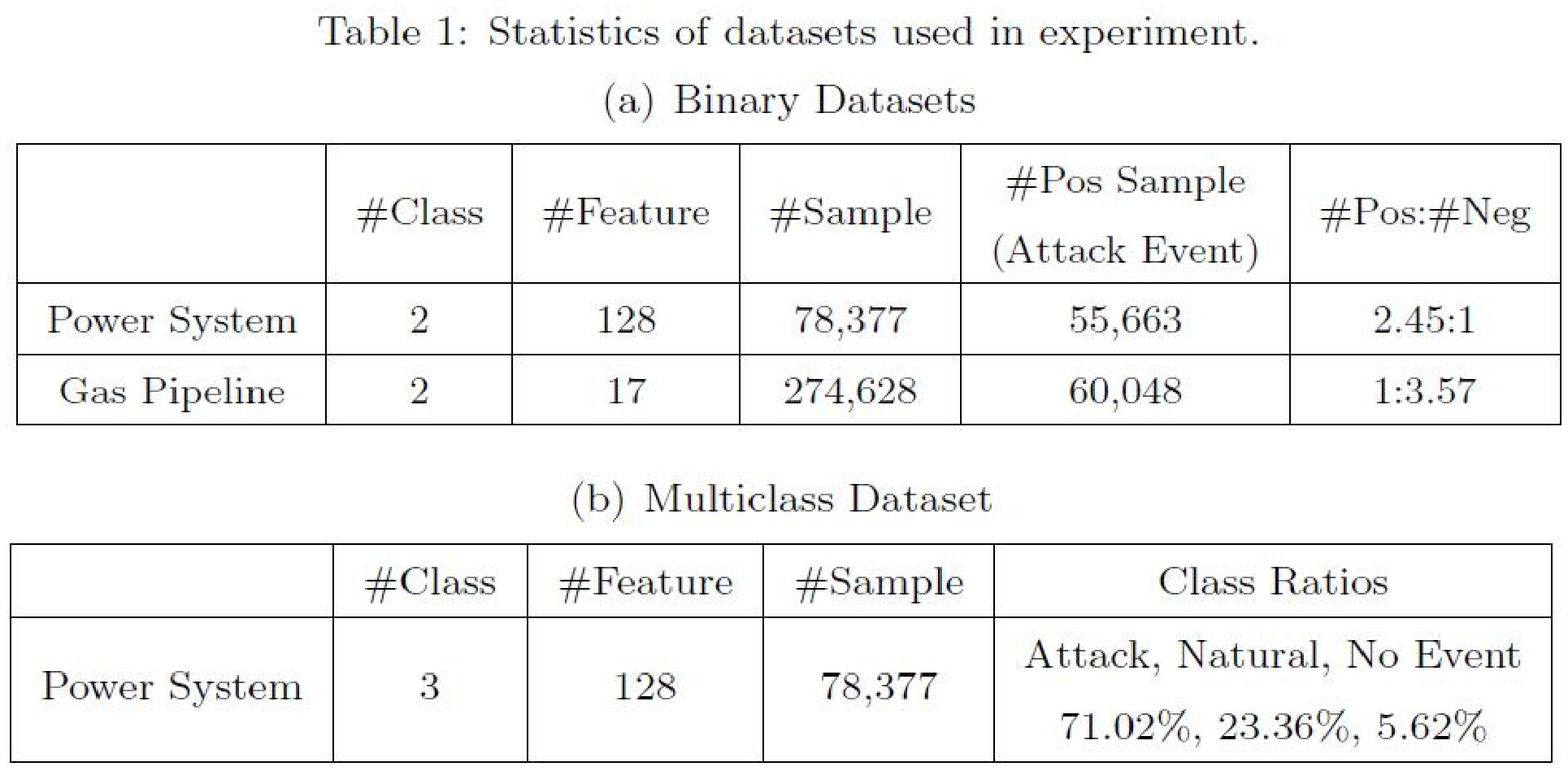}
\caption*{Table 1:Statistics of datasets used in experiment.}
\end{figure}

As shown in Table~1, a disproportionate number of classes exist in all datasets.
For the Power System dataset, the number of positive samples representing attack events is larger than that of normal events.
This deviates from the popular belief that attacks are rare in a system.
However, it is noteworthy that for this dataset, the portion of each class is determined by the testbed's creator during simulation.
In fact, regardless of the class outnumbers, the class imbalance problem is prevalent in real-world applications.
Further, the algorithms to mitigate imbalanced classes do not rely on the meaning of a specific class.
Therefore, we are confident that the experimental results on testbeds described herein are applicable to real-world industrial control applications.

\subsection{Evaluation Metrics}
\label{sec:4.2}

We adopt the following metrics to evaluate the performance of online learning algorithms for the intrusion detection task.

\paragraph{Cumulative error rate}
The cumulative error rate is the ratio of the number of mistakes made by an online learner over the number of samples received to date.
Despite its extensive usage in online learning studies, the cumulative error rate is inept to measure class-imbalanced datasets, as it ignores the different costs of misclassifying different classes.
In an extreme case, one can create a trivial classifier on a highly imbalanced dataset (i.e., blanket prediction of the majority class) that exhibits a low error rate but is in fact of little use.

\paragraph{Sensitivity}
Sensitivity, or true positive rate, measures the proportion of positives that are identified correctly as such (e.g., the percentage of attacks identified correctly by the intrusion detection system).
For a binary classification problem, let $P$ denote the number of positive samples, and $N$ the number of negative samples.
Further, let $TP$, $TN$, $FP$, and $FN$ denote the true positive, true negative, false positive, and false negative, respectively.
The sensitivity can be calculated as follows:
\begin{equation*}
sensitivity = \frac{TP}{P} = \frac{TP}{TP+FN}
\label{eq4.2-1}
\end{equation*}

\paragraph{Specificity}
Specificity, or true negative rate, measures the proportion of negatives that are identified correctly as such.
\begin{equation*}
specificity = \frac{TN}{N} = \frac{TN}{TN+FP}
\label{eq4.2-2}
\end{equation*}
It is noteworthy that the sensitivity and specificity, by their definitions, are only applicable to the binary classification test.

\paragraph{Weighted sum of sensitivity and specificity}
The weighted sum of sensitivity and specificity (abbreviated as ``sum'' hereafter) is defined as follows:
\begin{equation*}
sum = \eta_p \times sensitivity + \eta_n \times specificity
\label{eq4.2-3}
\end{equation*}
where $0 \leq \eta_p \leq 1$, $0 \leq \eta_n \leq 1$, and $\eta_p + \eta_n = 1$.
As a cost-sensitive metric, the weighted sum is suitable for measuring a classifier's performance on the class-imbalanced dataset.
The higher the sum value, the better the classifier.
When the $\eta_p$ and $\eta_n$ are both equal to 0.5, the sum becomes the well-known balanced accuracy.
In our experiment, we set $\eta_p$ and $\eta_n$ to 0.5.

\subsection{Benchmark Setup}
\label{sec:4.3}

The intent of our work is to establish a foundation for the application of online learning to intrusion detection in industrial control systems.
The benchmarks selected are thus state-of-the-art online learning algorithms with distinctive features.
We employ the SOL online learning library~\cite{DBLP:journals/ijon/WuHLLSY17} in our experiment for its good accessibility and efficiency.
As SOL includes a number of online learning methods, only representative methods are reported herein.

Specifically, the first-order online learning algorithms used in our experiment are as follows:
\begin{itemize}
\item Perceptron: the classical online learning algorithm~\cite{rosenblatt1958perceptron};
\item ALMA: the approximate large margin algorithm~\cite{DBLP:journals/jmlr/Gentile01};
\item ROMMA: the relaxed online maximum margin algorithm~\cite{DBLP:journals/ml/LiL02};
\item OGD: the online gradient descent algorithm~\cite{DBLP:conf/icml/Zinkevich03};
\item PA: the passive aggressive online learning algorithm~\cite{DBLP:journals/jmlr/CrammerDKSS06};
\item CSOGD: the cost sensitive online gradient descent algorithm~\cite{DBLP:conf/icdm/ZhaoZWLH15}.
\end{itemize}

The second-order online learning algorithms include the following:
\begin{itemize}
\item CW: the confidence-weighted learning algorithm~\cite{DBLP:conf/icml/DredzeCP08};
\item AROW: the adaptive regularization of weight vectors algorithm~\cite{DBLP:conf/nips/CrammerKD09};
\item SCW: the soft confidence weighted learning algorithm~\cite{DBLP:conf/icml/HoiWZ12};
\item ARCSOGD: the adaptive regularized cost-sensitive online gradient descent algorithm~\cite{DBLP:conf/icdm/ZhaoZWLH15};
\item ARCSMC: the adaptive regularized cost-sensitive multiclass online learning algorithm presented in Section~\ref{sec:3.5}.
\end{itemize}

The experiment is conducted on a PC with a 2.4-GHz CPU and 8-GB RAM. 
The key parameters for each algorithm are chosen from a small range of values (i.e., 0.001, 0.01, 0.1, 1, 10, 100, 1000) on a validation set.
The elements of the cost matrix described in Section~\ref{sec:3.5} are set inversely proportional to each class count.
We shuffled the sample order for each dataset randomly and repeated the experiment 10 times with new shuffles. The average results and corresponding
standard deviations over 10 trials are reported in Table 2.

\begin{figure}[tpb]
\centering 
\includegraphics[width=0.9\textwidth]{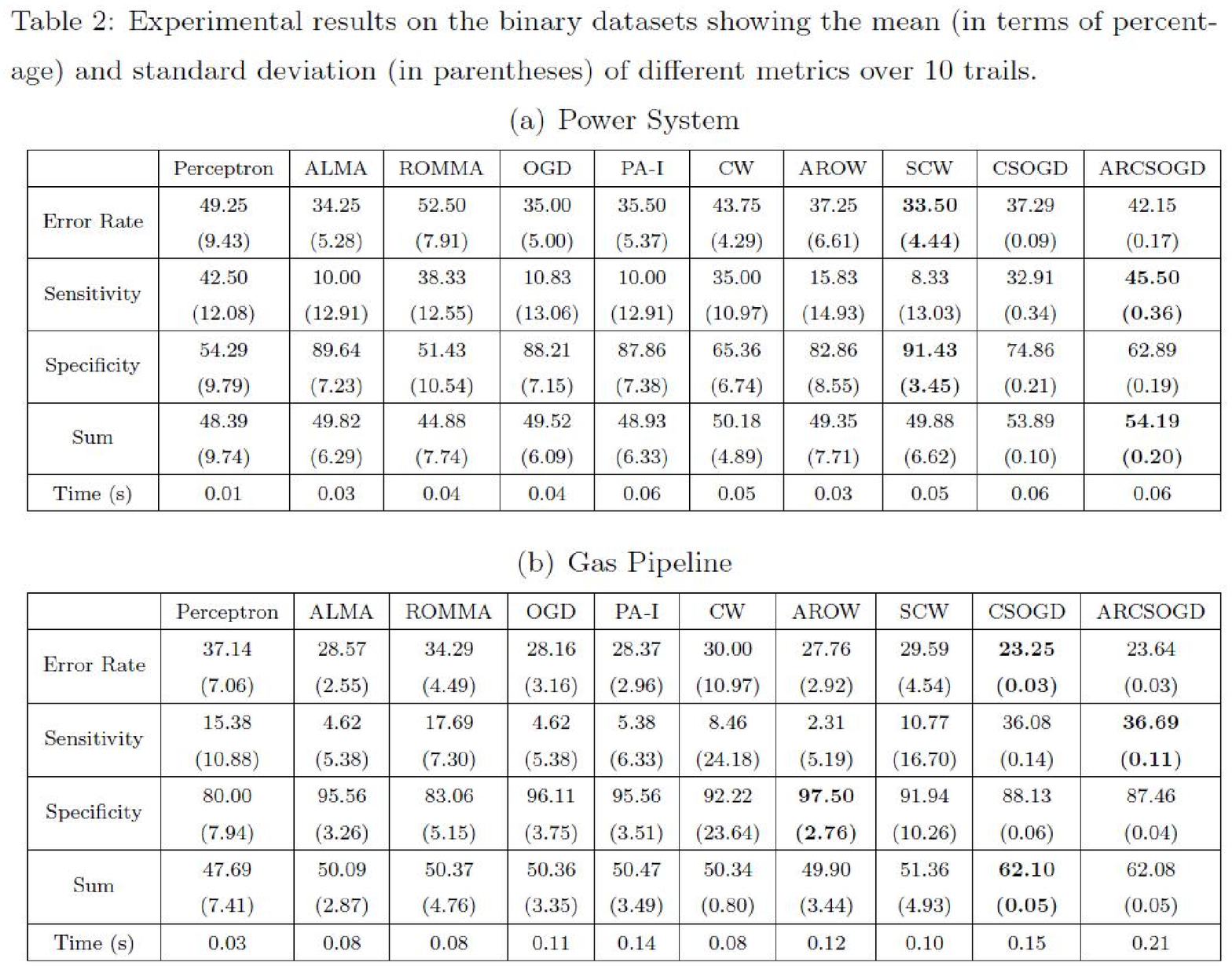}
\caption*{Table 2: Experimental results on the binary datasets showing the mean (in terms of percentage) and standard deviation (in parentheses) of different metrics over 10 trails.}
\end{figure}

\subsection{Evaluation of Binary Datasets}
\label{sec:4.4}

\begin{figure}[t]
\centering
\includegraphics[width=1.0\textwidth]{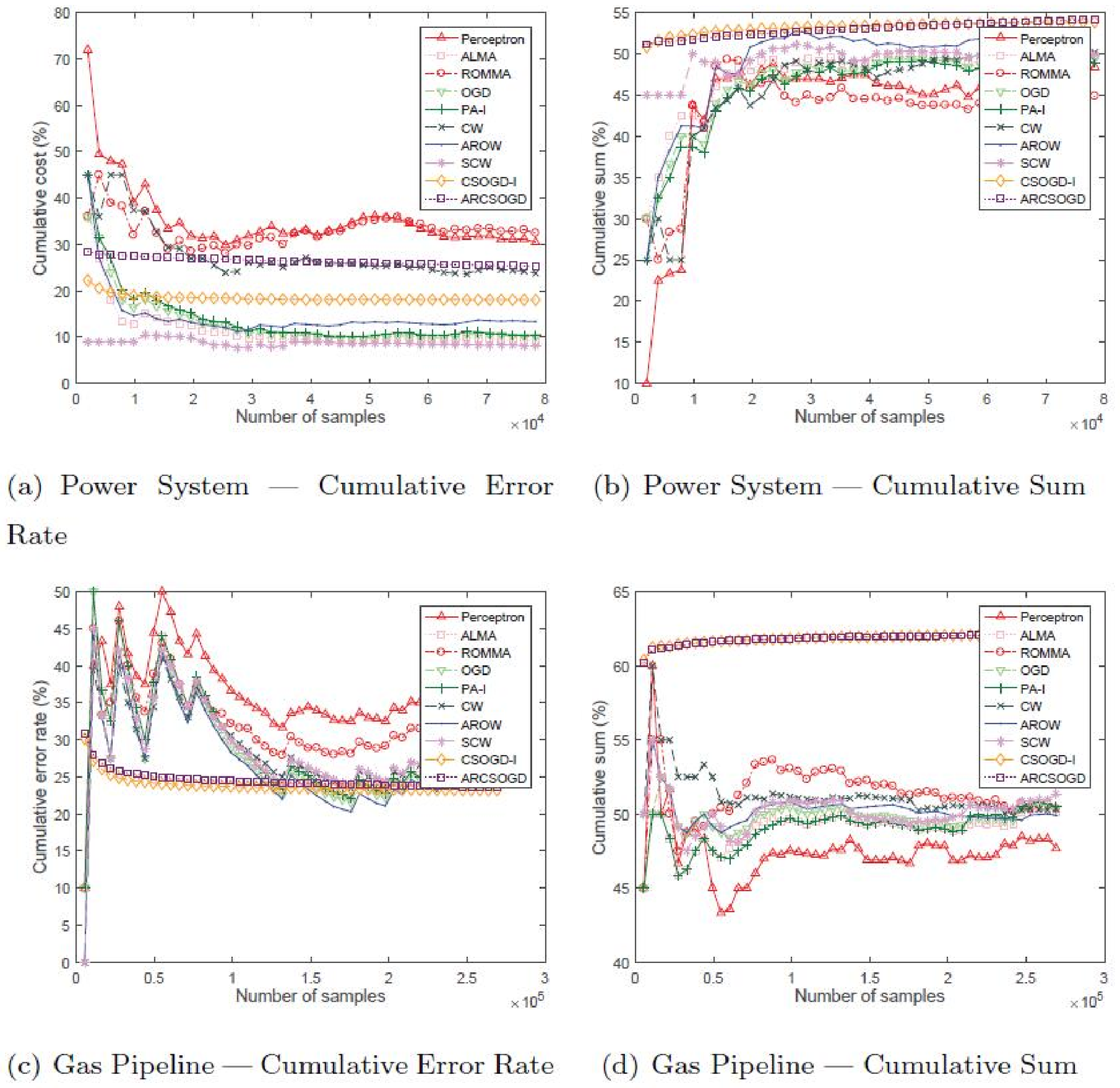}
\caption{The variation in evaluation metrics along the entire online learning process on the binary datasets.}
\label{fig4-1}
\end{figure}

Table~2 reports the mean error rate, sum, sensitivity, and specificity, together with their standard deviations of different algorithms measured at the last learning round on the two binary datasets.
Figure~\ref{fig4-1} depicts the variation in mean error rate and sum along the entire online learning process.
The running-time for each algorithm, i.e., the total time (in seconds) consumed by updating the models, and generating the predictions is included in Table~2 as well.
Our observations from these results are as follows.

First, the classification problem is complex as the error rates are high.
This has been confirmed previously~\cite{DBLP:conf/icmla/BeaverBB13,hink2014machine}, where the classifiers tend to make mistakes on the rare classes.
As for the online learning algorithms evaluated in this experiment, their error rates are approximately 30\% in most cases, not differing significantly from those of the batch models reported in~\cite{DBLP:conf/icmla/BeaverBB13,hink2014machine}.
Considering the minimal time cost of online learning algorithms (e.g., they can process hundred thousands of samples in tens of microseconds with moderate computing resource), we conclude that online learning can detect cyberattacks of industrial control systems more effectively compared to conventional batch learning approaches.

Next, the sensitivity records are less than those of specificity.
For example, the sensitivity against specificity of the ALMA algorithm is 4.62\% against 95.56\% on the Gas Pipeline dataset.
This is to be expected, as for a class-imbalanced dataset, identifying samples of the rare class (positive class in this case) is more difficult than that of the majority class (negative class).
A classifier trained under this setting is more likely to err on the positive samples, rendering the metric for measuring the correctly identified positive samples, i.e., the sensitivity record, low.
Because an intrusion detection system focuses more on the percentage of real attacks that are correctly identified, one should focus on promoting the performance in terms of sensitivity in practice.

Finally, by examining the values of sensitivity, specificity, and their weighted sum, we discovered that the two cost-sensitive online learners, i.e., CSOGD and ARCSOGD, outperform the others that do not apply the cost-sensitive learning schema in most cases.
This suggests that it is effective to apply cost-sensitive algorithms on the intrusion detection task for industrial control systems.

\subsection{Evaluation of Multiclass Datasets}
\label{sec:4.5}

We further evaluate the performance of online learning algorithms for multiclass classification in terms of their error rates and cost-sensitive metrics.
As metrics such as sensitivity and specificity are for the binary classification test, we adapted for the multiclass dataset by treating one class as positive and all other classes as negative.
We calculated the cost-sensitive metrics for each class as such, and reported them in Table~3 and Figure~\ref{fig4-2}.

\begin{figure}[tpb]
\centering 
\includegraphics[width=0.9\textwidth]{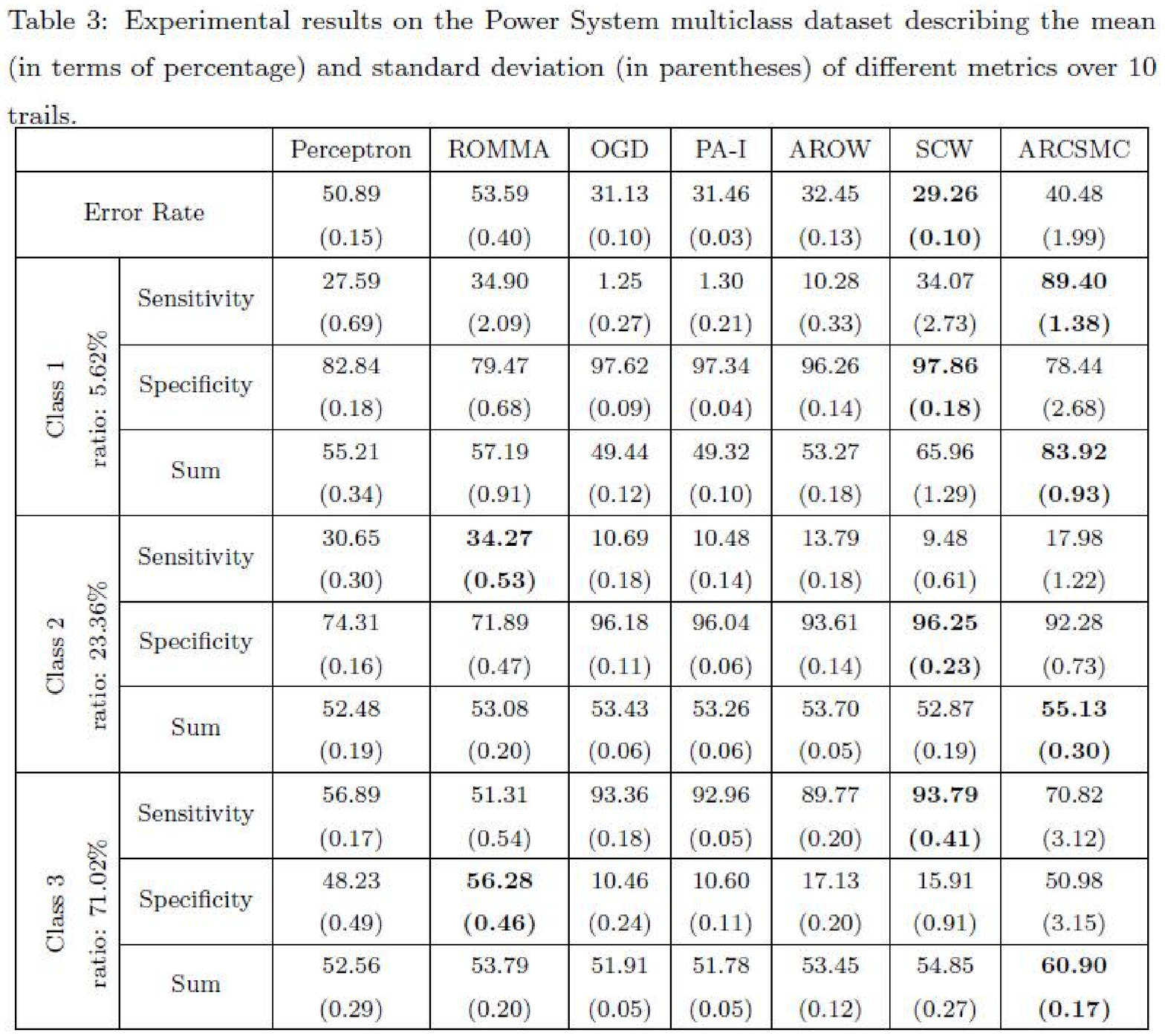}
\caption*{Table 3: Experimental results on the Power System multiclass dataset describing the mean
(in terms of percentage) and standard deviation (in parentheses) of different metrics over 10
trails..}
\end{figure}

Similar to the result of binary datasets, the cost-sensitive learning algorithm (ARCSMC) generally outperforms regular online learners in terms of the weighted sum of sensitivity and specificity.
This again indicates the necessity of applying cost-sensitiveness for the class-imbalanced problem. 
As shown, SCW achieves the lowest error rate, at the cost of low sensitivity values on two minority classes.
This indicates that SCW tends to classify all samples as the majority class.
From a practical standpoint, such a classifier is not very helpful though it demonstrates a low error rate.
Therefore, we conclude that ARCSMC is better than SCW when applied to detect the few but significant intrusion events in industrial control systems.

\section{Conclusion}
\label{sec5}

We herein explored the viability of applying online learning algorithms to perform intrusion detection in industrial control systems.
We began by a brief review of the industrial control systems, the cyberthreats they experienced, and the intrusion detection methods, especially those based on machine learning techniques.
We highlighted that because the industrial control systems require real-time response and uninterrupted operations, any algorithms that they employ to detect attacks should be efficient, scalable, and suitable for processing data stream.
Online learning algorithms satisfy only these requirements.
We subsequently introduced several state-of-the-art online learning methods, and especially the cost-sensitive online classification that could be used to improve the prediction accuracy of the rare but significant attack events.
We applied the cost-sensitive learning scheme to AROW~\cite{DBLP:conf/nips/CrammerKD09} to derive a new method---the adaptive regularized cost-sensitive multiclass online learning (ARCSMC).
The experimental results indicated that the cost-sensitive online learning algorithms, in particular the proposed ARCSMC, are both effective and efficient for detecting cyberattacks in industrial control systems.
\begin{figure}[tpb]
\centering
\includegraphics[width=1\textwidth]{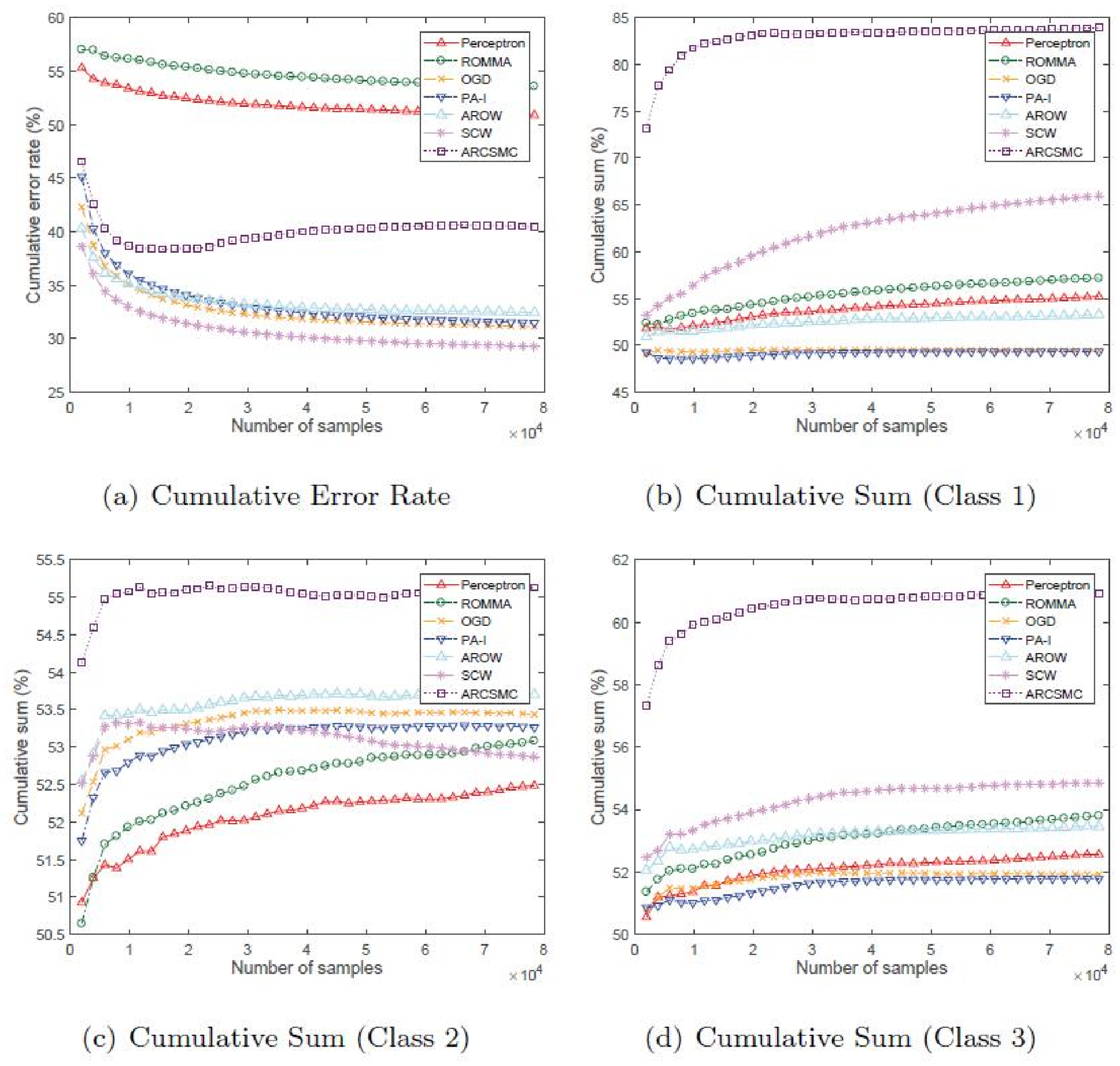}
\caption{The variation of evaluation metrics along the entire online learning process on the Power System multiclass datasets.}
\label{fig4-2}
\end{figure}

For future work, we wish to extend our experiments to a more substantial size dataset and to more applications.
This involves building new testbeds to mimic more industrial control processes and simulating more cyberattacks with different communication protocols and attack schemes.
In addition, more online learning algorithms and classification schemes will be studied.
One possible solution is the online one-class classification that is particularly suitable for problems where the majority of available data represents the normal behavior of the system, whereas the data related to attack events are difficult to obtain.
In conclusion, our work serves as an initial attempt at applying online learning to detect cyberattacks in industrial control systems.

\end{document}